\documentclass[conference]{IEEEtran}
\renewcommand{\footnoterule}{%
    \kern -3pt 
    \hrule width 0.4\linewidth height 0.4pt 
    \kern 2pt 
}
\IEEEoverridecommandlockouts
\usepackage{cite}
\usepackage{amsmath,amssymb,amsfonts}
\usepackage{algorithmic}
\usepackage{algorithm}
\usepackage{graphicx}
\usepackage{textcomp}
\usepackage{booktabs}
\usepackage[table]{xcolor}
\def\BibTeX{{\rm B\kern-.05em{\sc i\kern-.025em b}\kern-.08em
    T\kern-.1667em\lower.7ex\hbox{E}\kern-.125emX}}
\begin{document}

\title{A Knowledge Noise Mitigation Framework for Knowledge-based Visual Question Answering}

\author{

\IEEEauthorblockN{Zhiyue Liu$^{1,2*}$, Sihang Liu$^{1}$, Jinyuan Liu$^{1}$, Xinru Zhang$^{1}$}
\IEEEauthorblockA{$^1$School of Computer, Electronics and Information, Guangxi University, Nanning, China}
\IEEEauthorblockA{$^2$Guangxi Key Laboratory of Multimedia Communications and Network Technology}
\IEEEauthorblockA{liuzhy@gxu.edu.cn, \{2313301037, 2213394017, 2313301061\}@st.gxu.edu.cn}}

\maketitle

\renewcommand{\thefootnote}{}
\footnotetext{$^*$Corresponding author.}

\begin{abstract}
Knowledge-based visual question answering (KB-VQA) requires a model to understand images and utilize external knowledge to provide accurate answers. Existing approaches often directly augment models with retrieved information from knowledge sources while ignoring substantial knowledge redundancy, which introduces noise into the answering process. To address this, we propose a training-free framework with knowledge focusing for KB-VQA, that mitigates the impact of noise by enhancing knowledge relevance and reducing redundancy. First, for knowledge retrieval, our framework concludes essential parts from the image-question pairs, creating low-noise queries that enhance the retrieval of highly relevant knowledge. Considering that redundancy still persists in the retrieved knowledge, we then prompt large models to identify and extract answer-beneficial segments from knowledge. In addition, we introduce a selective knowledge integration strategy, allowing the model to incorporate knowledge only when it lacks confidence in answering the question, thereby mitigating the influence of redundant information. Our framework enables the acquisition of accurate and critical knowledge, and extensive experiments demonstrate that it outperforms state-of-the-art methods.
\end{abstract}


\section{Introduction}
\label{sec:intro}

Knowledge-based visual question answering (KB-VQA)~\cite{marino2019ok} is an extension of the traditional VQA task~\cite{antol2015vqa} by requiring models to answer questions that cannot be resolved solely from image content, necessitating the integration of external knowledge. Early methods~\cite{marino2021krisp,luo2021weakly} focus on extracting explicit knowledge from multiple open-domain knowledge resources (e.g., ConceptNet~\cite{liu2004conceptnet}, Wikipedia~\cite{vrandevcic2014wikidata}, and Google Search Corpus~\cite{luo2021weakly}) and combining the retrieved knowledge with image questions for answering. However, retrieving knowledge relevant to both images and questions presents challenges, often leading to the introduction of noise. Inspired by the powerful knowledge reasoning capabilities of large language models (LLMs), subsequent studies~\cite{yang2022empirical,hu2023promptcap,shao2023prompting,xenos2023simple} prompt LLMs with textual descriptions converted from images to generate answer predictions with no need to retrieve explicit knowledge.

\begin{figure}[t]
\centering
\centerline{\includegraphics[width=0.50\textwidth]{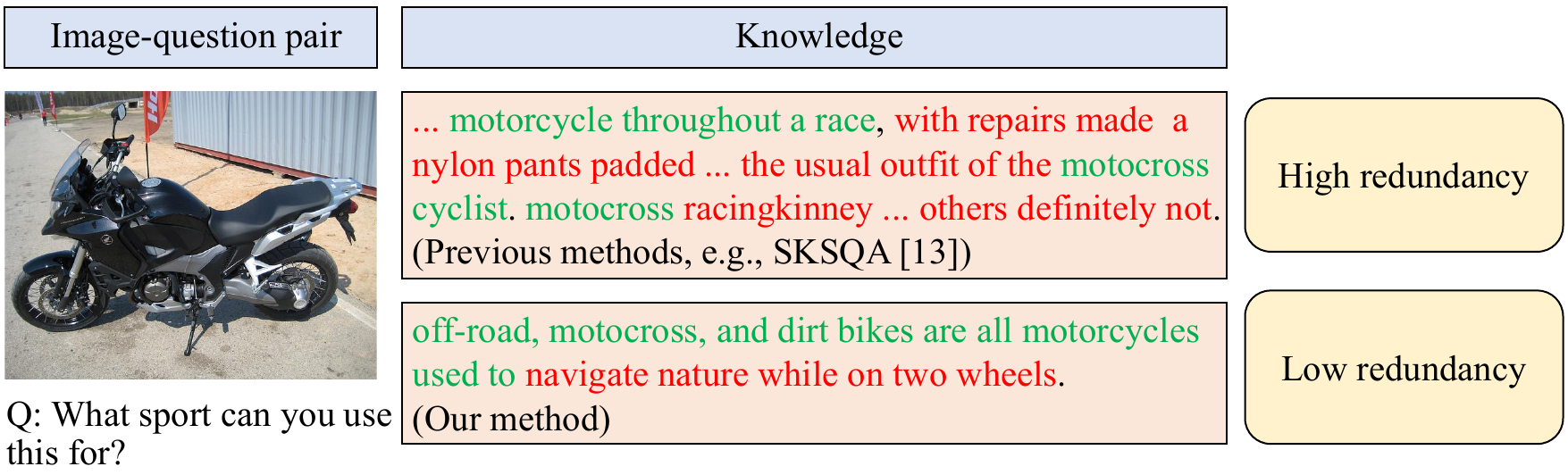}}
\caption{External knowledge retrieved by our method and recent works. In the knowledge texts, words with low relevance to the image-question pair are highlighted in red, while relevant words are highlighted in green.}
\label{fig1}
\end{figure}

The limited knowledge scope and lack of visual perception challenge LLMs, while visual language models (VLMs) are more suited for vision-centric tasks than knowledge-centric ones~\cite{an2024knowledge}. Therefore, another line of work focuses on how to integrate external knowledge bases on smaller LLMs for KB-VQA. \cite{lin2022retrieval} translates image content into textual semantics, which consists of an image description, objects with attributes, and OCR characters. Then, a long query containing these texts is built to retrieve knowledge that assists in LLM training. However, their retrieved knowledge often includes a substantial amount of content unrelated to the question, which may limit the model's performance. Recent work~\cite{hao2024self} explores training VLMs to pick up relevant knowledge documents from retrieved information, while many question-unrelated segments exist in those documents, as in Fig.~\ref{fig1}. Moreover, the lengthy search queries constructed by existing works make it difficult for the retrieval process to focus on key information in image-question pairs, often resulting in a large amount of redundant content. Inaccurate knowledge introduces additional burdens, hindering the answering process.

To mitigate the impact of knowledge noise, we propose a training-free framework with knowledge focusing for KB-VQA (KF-VQA for short), which enhances knowledge relevance and reduces redundancy. Firstly, for knowledge retrieval, we aim to construct low-noise queries, thus increasing the relevance of selected knowledge. Since the image provides key context for identifying relevant knowledge, we leverage the multimodal perception capabilities of VLMs to distill the essential content from the image-question pair. This enables us to obtain precise and low-noise queries, minimizing the impact of irrelevant content in queries for knowledge retrieval. Secondly, considering that there is still redundant information in the retrieved knowledge documents, we refine the knowledge by integrating the VLM's visual understanding with the LLM's reasoning capabilities to extract the relevant segments. That is, LLMs combine the question with fine-grained visual details extracted by VLMs to identify the answer-beneficial content, thereby reducing knowledge redundancy. However, completely filtering out noisy content from external knowledge is impractical~\cite{hao2024self}. It is intuitive to lower the impact of redundant knowledge by introducing external knowledge only when the model requires it for question answering. Therefore, we introduce a selective knowledge integration strategy. LLMs incorporate external knowledge to enhance the reasoning ability only when they lack confidence in answering a question, further reducing the impact of knowledge noise. Experiments on benchmark datasets show that KF-VQA outperforms baselines and achieves the state-of-the-art performance.

Our contributions are summarized as follows:

\begin{itemize}
\item We propose a training-free framework with knowledge focusing for KB-VQA that mitigates the noise effects by enhancing knowledge relevance and reducing redundancy.
\item We introduce a selective knowledge integration strategy which incorporates knowledge to help answer questions when LLMs lack confidence in answering, further reducing the impact of redundant information.
\item Experimental results show that our method outperforms competitive baselines, achieving the state-of-the-art results on both OK-VQA~\cite{marino2019ok} and A-OKVQA~\cite{schwenk2022okvqa} datasets.

\end{itemize}

\section{Related Work}

\subsection{Training-based KB-VQA Methods}
Training-based KB-VQA methods require training a model to answer questions. Early studies~\cite{marino2021krisp,luo2021weakly} augment models by retrieving knowledge directly from various open-domain world knowledge sources. Given the powerful knowledge inference capabilities of LLMs, subsequent methods explore how to combine explicit knowledge with LLMs implicit knowledge. For example, \cite{gui2022kat, lin2022revive} generate candidate answers to questions by prompting LLMs and incorporate the retrieved knowledge to enhance model performance, while ~\cite{lin2022retrieval, lin2023fine} utilize explicit external knowledge to train LLMs. Since irrelevant information exists in the retrieved knowledge, \cite{hao2024self} trains VLMs to identify relevant document-level knowledge. However, within the documents, there is still a large amount of content irrelevant to the questions. Our framework could identify fine-grained knowledge segments related to the image-question pair through the collaboration of VLMs and LLMs, all without the need for training.

\subsection{Training-free KB-VQA Methods}
Although training-based methods achieve performance improvement, they often rely heavily on large amounts of labeled data and suffer from poor generalization. With the emergence of knowledge reasoning capabilities in LLMs, a training-free method~\cite{yang2022empirical} prompts GPT-3 with in-context examples and image captions to realize KB-VQA, while generic captions may lose image details. \cite{hu2023promptcap} assists GPT-3 in understanding images with question-guided captions. \cite{shao2023prompting} selects better in-context examples to maximize the potential of GPT-3.  Given the high cost of GPT-3 API calls, \cite{xenos2023simple} constructs a large amount of in-context content to achieve GPT-3-like performance using smaller and open-source LLMs. Due to the limited knowledge scope of LLMs and the flaw of hallucination, \cite{an2024knowledge} decomposes questions to retrieve knowledge from external knowledge bases, and shows that VLMs fall short in KB-VQA. However, this work still relies on lengthy queries to search for knowledge, which leads to the retrieval of irrelevant content. In contrast, our KF-VQA not only uses essential content from the image-question pair as low-noise queries, but also filters out irrelevant content from retrieved knowledge, thereby enhancing the knowledge relevance.

\begin{figure*}[!t]
\centering
\centerline{\includegraphics[width=0.74\textwidth]{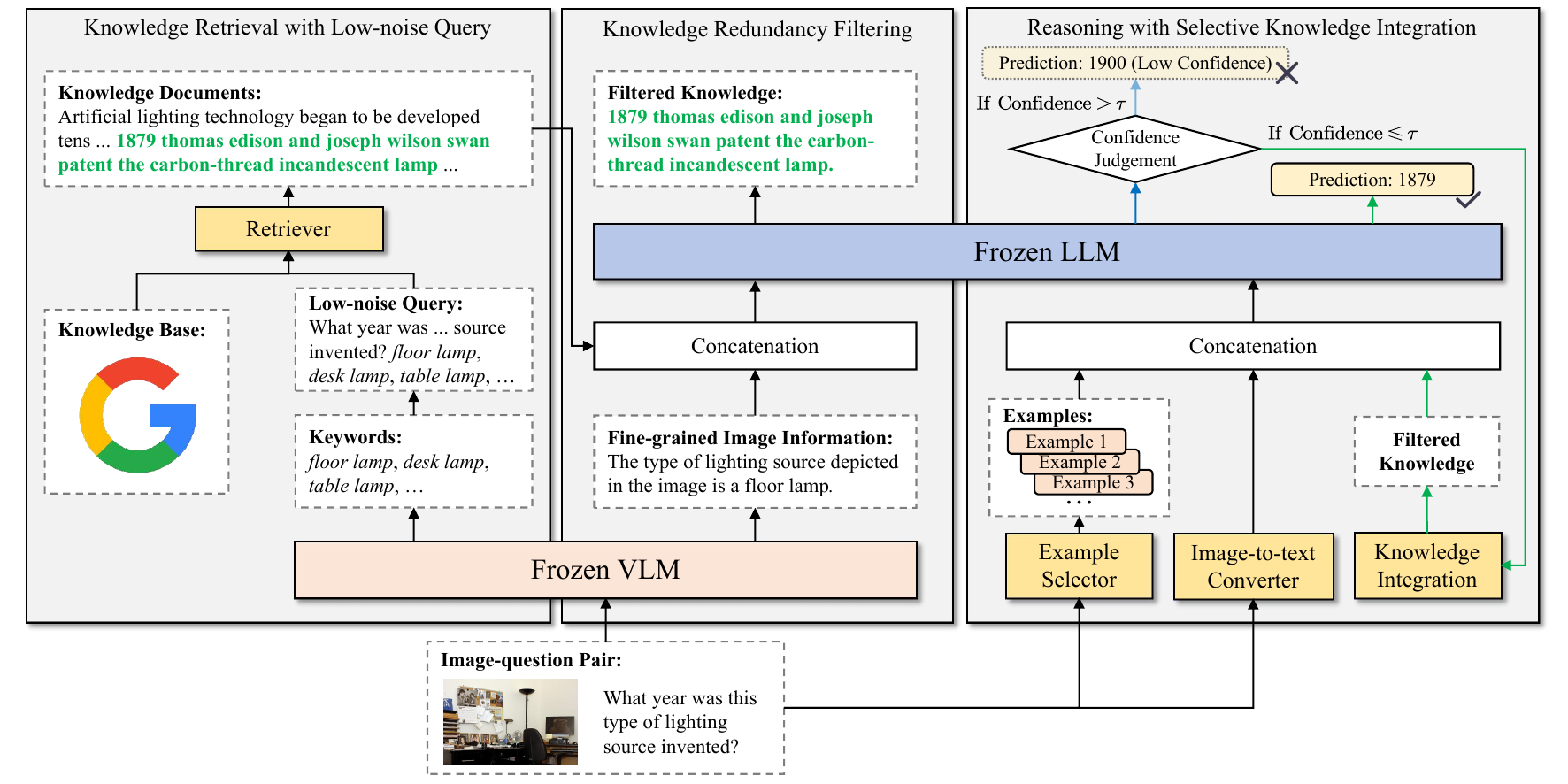}}
\caption{Overview of the proposed framework. KF-VQA constructs low-noise queries by extracting essential content from the image and question to retrieve relevant knowledge documents. Then, a redundancy filtering process extracts fine-grained segments from the retrieved documents. During LLM reasoning, knowledge integration is selectively activated (green line) only when the prediction confidence without knowledge (blue line) is low.}
\label{fig2}
\end{figure*}

\section{Methodology}

The overall architecture of KF-VQA is illustrated in Fig.~\ref{fig2}. Designed to enhance knowledge relevance and reduce redundancy, it comprises three components: knowledge retrieval with low-noise queries, knowledge redundancy filtering, and reasoning with selective knowledge integration. To retrieve highly relevant knowledge, the retrieval module builds low-noise queries based on essential content extracted from image-question pairs. The filtering module then leverages the complementary strengths of LLMs and VLMs to select fine-grained segments from the retrieved information. During LLM reasoning, a selective knowledge integration strategy incorporates knowledge segments only when the model lacks confidence in its answer, thereby minimizing the impact of noise.

\subsection{Knowledge Retrieval with Low-noise Query}
KB-VQA typically involves retrieving auxiliary information from knowledge bases. However, the lengthy retrieval queries, containing excessive image details, from existing works~\cite{lin2022retrieval,hao2024self} always distract the retrieval process, resulting in the selection of irrelevant content. We propose constructing low-noise queries containing only essential content from image-question pairs, guiding the retriever to focus on key information and select highly relevant knowledge. Specifically, we utilize the multimodal perception capability of VLMs to extract keywords that summarize an image-question pair. That is, given an image-question pair $(v, q)$, we prompt a frozen VLM to generate a set of concise keywords $k$ which represent the pair's essential content:
\begin{equation}
k={\rm VLM}(\mathcal{P}_1,v,q), \label{eq2}
\end{equation}
where $\mathcal{P}_1$ is the prompt template. Considering the importance of the question in knowledge retrieval, we concatenate the keywords $k$ with the question $q$ to form a low-noise query $q_l = [q;k]$. Then, we use the Google Search Corpus~\cite{luo2021weakly} as our knowledge base $\mathcal{D}$ and adopt $q_l$ to find relevant knowledge documents from it. Following~\cite{lin2022retrieval, hao2024self}, we encode $q_l$ and each knowledge document $d_i$ from $\mathcal{D}$, and calculate the similarity score between $q_l$ and $d_i$ to retrieve the top-$r$ documents with the highest similarity as follows:
\begin{equation}
K_d = \{\hat{d}_1, \dots, \hat{d}_r\} = {\rm Top}_r({\rm sim}(q_l, d_1),{\rm sim}(q_l, d_2),\dots),
\label{eq:top_r}
\end{equation}
where ${\rm sim}(q_l, d_i)$ denotes the dot product between the embeddings of $q_l$ and $d_i$. Since $q_l$ contains only key content, the retrieval process would focus on these critical pieces to find relevant documents $K_d$.

\subsection{Knowledge Redundancy Filtering}
Although the retrieval with low-noise queries could enhance the relevance of retrieved knowledge documents to the image-question pair, considerable redundancy remains in these documents, impeding model reasoning. We guide LLMs to filter redundant segments from retrieved documents by leveraging question-relevant visual details, since focusing on the visual aspects of questions helps identify cues for selecting relevant segments. To extract such visual details, we first prompt the frozen LLM to transform the original question $q$ into an image-based question $q_v$, which focuses exclusively on the content presented in the image $v$. Then, the VLM leverages its strong visual perception capabilities to answer this visual perception question $q_v$ to obtain required image information as follows:
\begin{equation}
q_v={\rm LLM}(\mathcal{P}_2,q), \label{eq5}
\end{equation}
\begin{equation}
a_v={\rm VLM}(v,q_v), \label{eq6}
\end{equation}
where $\mathcal{P}_2$ is the prompt template for LLMs, and $a_v$ denotes the fine-grained visual details about the question $q$.

Based on the question $q$ and visual details $a_v$, we encourage the LLM to identify relevant segments $K_s$ from $K_d$ as follows:
\begin{equation}
K_s={\rm LLM}(\mathcal{P}_3,K_d,q,a_v), \label{eq7}
\end{equation}
where $\mathcal{P}_3$ is the prompt template. The segments $K_s$ provide more fine-grained, low-redundancy knowledge about the image-question pair compared to the documents $K_d$, thereby establishing a better reasoning basis for the LLM.

\subsection{Reasoning with Selective Knowledge Integration}
During the reasoning process, our knowledge segments $K_s$ would assist the LLM in making predictions. However, despite our efforts to filter out redundancies, completely eliminating knowledge noise is impractical. For LLMs, there are situations where the model relies on its implicit knowledge to answer a question correctly, while the retrieved knowledge leads to an incorrect answer~\cite{hao2024self}. Inspired by this, we propose a selective knowledge integration strategy to minimize the impact of residual noise. This strategy aligns with human intuition and enables the model to integrate external knowledge to improve its reasoning only when it is uncertain about its response. 

Measuring prediction certainty is central to our strategy. The generation probability assigned to an output sequence reflects the LLM confidence in its prediction. Therefore, we explore the model's confidence score, derived from probabilities, as a decision metric to assess whether knowledge is needed. Specifically, given the input $x$ and output $y = \{y^1,\dots,y^L\}$, we aggregate the sampling probabilities at each time step to obtain the confidence score as follows:
\begin{equation}
f(y) = \sum_{t=1}^{L} \log p_{\rm LLM}(y^t \mid x, y^{<t}), \quad s = e^{f(y)},
\end{equation}
where the confidence score $s$ is defined as the mapping of the aggregated probability $f(y)$ to a value between 0 and 1 using an exponential function. Then, we could achieve selective knowledge integration in LLMs by setting a confidence threshold $\tau$. The LLM would answer the question using its implicit knowledge and obtain a score $s$. If $s \leq \tau$, the LLM would introduce $h$ segments $K^h_s$ which are sampled from our retrieved knowledge $K_s$ to boost its reasoning.

Besides, in-context examples and generic image captions are essential for training-free methods. Following~\cite{xenos2023simple,an2024knowledge}, we use MCAN~\cite{yu2019deep} to select the top-$n$ most similar examples $E$ for each test sample and obtain image captions $c_v$ through a converter~\cite{tiong2022plug}. We combine $c_v$ with image details $a_v$ as $I$ to provide LLMs with visual semantics. Finally, the reasoning process with selective knowledge integration is obtained as:
\begin{equation}
a_f = 
\begin{cases} 
\underset{j \in \{1, \dots, m\}}{\arg\max} p_{\rm LLM}(a_j \mid \mathcal{P}_4, E_j, I, K^h_s, q), & \text{if } s \leq \tau, \\
\underset{j \in \{1, \dots, m\}}{\arg\max} p_{\rm LLM}(a_j \mid \mathcal{P}_4, E_j, I, q), & \text{otherwise},
\end{cases}
\end{equation}
where $\mathcal{P}_4$ is the prompt, and $s = e^{\underset{j}{\max}\  p_{\rm LLM}(a_j \mid \mathcal{P}_4, E_j, I, q)}$. To obtain more robust performance~\cite{xenos2023simple,yang2022empirical}, we prompt LLMs $m$ times with various $E_j$ to obtain $m$ predictions, and ensemble them to select the one with the highest probability.

\section{Experiments}

\subsection{Experimental Setup}

\textbf{Dataset.} We conduct experiments on two datasets: OK-VQA~\cite{marino2019ok} and A-OKVQA~\cite{schwenk2022okvqa} to evaluate the effectiveness of our method. OK-VQA is a challenging knowledge-based visual question answering dataset including 5K image-question pairs for testing, with most questions requiring external knowledge to answer. A-OKVQA, as an enhanced successor to OK-VQA, contains questions that not only demand a broader range of knowledge but also emphasize the use of knowledge for reasoning. It includes 1K image-question pairs for validation and 6K pairs for testing.

\textbf{Implementation Details.} We evaluate model performance using the standard VQA metric: accuracy~\cite{antol2015vqa}. In our framework, we employ Contriever~\cite{kamath2022webly} as the retriever, Llama3.2-11B-Vision~\cite{dubey2024llama} as the VLM, and Llama3-8B~\cite{dubey2024llama} as the LLM. We set the number of retrieved knowledge items $r$ to 20, the number of selected knowledge segments $h$ to 7, and the confidence threshold $\tau$ to 0.8. Following previous methods~\cite{xenos2023simple,an2024knowledge}, hyper-parameters $m$ and $n$ are set to 5 and 10, respectively.

\textbf{Baselines.}
We compare our framework with training-based and training-free methods. Training-based methods include KRISP~\cite{marino2021krisp}, Vis-DPR~\cite{luo2021weakly}, REVIVE~\cite{lin2022revive}, RA-VQA-v2~\cite{lin2023fine}, and SKSQA~\cite{hao2024self}. Training-free methods include PICa~\cite{yang2022empirical}, PromptCap~\cite{hu2023promptcap},  Prophet~\cite{shao2023prompting}, Simple~\cite{xenos2023simple}, and DKA~\cite{an2024knowledge}.

\subsection{Experimental Results}

\textbf{Comparison Results on OK-VQA.} Table~\ref{tab1} compares the performance of our KF-VQA with state-of-the-art baselines on the OK-VQA dataset. The proposed framework consistently outperforms both training-based and training-free methods, showing that our high-quality knowledge benefits LLM in knowledge-centric visual question answering. Training-based baselines select excessive knowledge using verbose queries, which hinders their performance due to knowledge redundancy. Although SKSQA employs document-level knowledge filtering, substantial irrelevant segments still exist within chosen documents. In contrast, without expensive training, our framework leverages low-noise queries and fine-grained knowledge selection to obtain concise and relevant knowledge segments. Compared to the strongest training-free baseline, DKA, KF-VQA could improve accuracy by 1.1\%. DKA also utilizes lengthy queries and struggles to retrieve relevant knowledge. The redundancy in DKA's knowledge prevents its LLMs from performing well, while our method integrates VLMs and LLMs to refine knowledge, thereby achieving the best performance. Besides, our concise knowledge reduces the length of LLM prompts, which accelerates the inference speed.

\begin{table}[t]
\centering
\caption{Comparison results on the OK-VQA dataset. The best result is in bold.}
\resizebox{1.01\linewidth}{!}{
\begin{tabular}{lccc}
\toprule
\textbf{Method} & \textbf{Model} & \textbf{Knowledge Resource} & \textbf{Acc (\%)} \\
\midrule
\multicolumn{4}{c}{\textcolor{gray}{\textbf{Training-based methods}}} \\ 
\textcolor{gray}{KRISP\cite{marino2021krisp}} & \textcolor{gray}{MMBERT} & \textcolor{gray}{ConceptNet+Wikipedia} & \textcolor{gray}{38.4} \\
\textcolor{gray}{Vis-DPR\cite{luo2021weakly}} & \textcolor{gray}{LXMERT} & \textcolor{gray}{Google Search} & \textcolor{gray}{39.2} \\ 
\textcolor{gray}{REVIVE\cite{lin2022revive}} & \textcolor{gray}{T5-large} & \textcolor{gray}{Wikipedia+GPT-3 (175B)} & \textcolor{gray}{58.0} \\ 
\textcolor{gray}{RA-VQA-v2\cite{lin2023fine}} & \textcolor{gray}{BLIP2 T5-XL (3B)} & \textcolor{gray}{Google Search} & \textcolor{gray}{62.1} \\ 
\textcolor{gray}{SKSQA\cite{hao2024self}} & \textcolor{gray}{BLIP2 T5-XL (3B)} & \textcolor{gray}{Google Search} & \textcolor{gray}{62.8} \\
\midrule
\multicolumn{4}{c}{\textbf{Training-free methods}} \\ 
PICa\cite{yang2022empirical} & GPT-3 (175B) & - & 48.0 \\ 
PromptCap\cite{hu2023promptcap} & GPT-3 (175B) & - & 60.4 \\ 
Prophet\cite{shao2023prompting} & GPT-3 (175B) & - & 61.1 \\ 
Simple\cite{xenos2023simple} & LLaMA 2 (13B) & - & 61.2 \\
DKA\cite{an2024knowledge} & LLaMA 2 (13B) & Wikipedia+ChatGPT & 62.1 \\ 
\midrule
KF-VQA (Ours) & LLaMA 2 (13B) & Google Search & 62.8 \\ 
KF-VQA (Ours) & LLaMA 3 (8B) & Google Search & \textbf{63.2} \\
\bottomrule
\end{tabular}
}
\label{tab1}
\end{table}

\begin{table}[t]
\centering
\caption{Comparison results on the A-OKVQA dataset.}
\begin{tabular}{lccc}
\toprule
\textbf{Method} & \textbf{Model} & \textbf{Val Acc} & \textbf{Test Acc}\\ 
\midrule
\multicolumn{4}{c}{\textcolor{gray}{\textbf{Training-based methods}}} \\ 
\textcolor{gray}{KRISP\cite{marino2021krisp}} & \textcolor{gray}{MMBERT} & \textcolor{gray}{33.7} & \textcolor{gray}{27.1}\\
\textcolor{gray}{SKSQA\cite{hao2024self}} & \textcolor{gray}{BLIP2 T5-XL (3B)} & \textcolor{gray}{57.2} & \textcolor{gray}{56.4}\\ 
\midrule
\multicolumn{4}{c}{\textbf{Training-free methods}} \\ 
PromptCap\cite{hu2023promptcap} & GPT-3 (175B) & 56.3 & 59.6 \\
Prophet\cite{shao2023prompting} & GPT-3 (175B) & 58.2 & 55.7 \\ 
Simple\cite{xenos2023simple} & LLaMA 2 (13B) & 58.6 & 57.5 \\ 
DKA\cite{an2024knowledge} & LLaMA 2 (13B) & 62.1 & 59.9 \\ 
\midrule
KF-VQA & LLaMA 2 (13B) & 62.4 & 60.5 \\ 
KF-VQA & LLaMA 3 (8B) & \textbf{62.9} & \textbf{60.9} \\
\bottomrule
\end{tabular}
\label{tab2}
\end{table}

\textbf{Comparison Results on A-OKVQA.} Table~\ref{tab2} presents the results on the validation and test sets of A-OKVQA.
Most training-based methods perform worse than training-free ones, while KF-VQA surpasses the best results of training-based methods by a large margin, achieving a 5.7\% increase on the validation set and a 4.5\% increase on the test set. The trained models focus on patterns within the training set, which may not generalize effectively to other questions of A-OKVQA that require diverse knowledge, and their reasoning ability is also inferior to that of strong LLMs. In contrast, our model encourages the LLM to perform accurate reasoning through concise knowledge. Compared to DKA, our method shows improvements of 0.8\% on the validation set and 1\% on the test set. That is, LLMs could better derive reasoning insights from our concise knowledge rather than redundant information.

\begin{table}[t]
\centering
\caption{Ablation study on OK-VQA.}
\begin{tabular}{lc}
\toprule
\textbf{Method} & \textbf{Acc}\\ 
\midrule
LLM & 61.3\\
\midrule
\emph{Baseline} & 60.4\\
\emph{Baseline} w/ LNQ & 61.5\\
\emph{Baseline} w/ LNQ \& KRF & 62.4\\
\emph{Baseline} w/ LNQ \& KRF \& SKI (KF-VQA) & \textbf{63.2}\\
\bottomrule
\end{tabular}
\label{tab3}
\end{table}

\subsection{Ablation Study}
To illustrate the effectiveness of the framework’s key components, we perform an ablation study on OK-VQA, with results shown in Table~\ref{tab3}. We build a \emph{Baseline} model which directly uses knowledge retrieved through lengthy
queries~\cite{lin2022retrieval,hao2024self} for reasoning. Baseline is inferior to the LLM without any external knowledge, as redundant and inaccurate knowledge fails to benefit training-free reasoning. We introduce low-noise queries for knowledge retrieval (i.e., w/ LNQ) and improve the performance of LLMs. That is, the queries with key information are beneficial for locating accurate knowledge. We further integrate knowledge redundancy filtering into knowledge processing (i.e., w/ LNQ \& KRF), resulting in improved performance that surpasses DKA. It shows that selecting relevant knowledge segments leads to fine-grained knowledge that is more accurate and concise. When the selective knowledge integration strategy (i.e.,  w/ LNQ \& KRF \& SKI) is applied, we obtain the complete framework (KF-VQA) which performs best. It shows that using knowledge without considering whether LLMs require it hinders performance, as noise from external information cannot be entirely filtered out. Each component contributes to the reasoning of LLMs.

\begin{figure}[t]
\centering
\centerline{\includegraphics[width=0.5\textwidth]{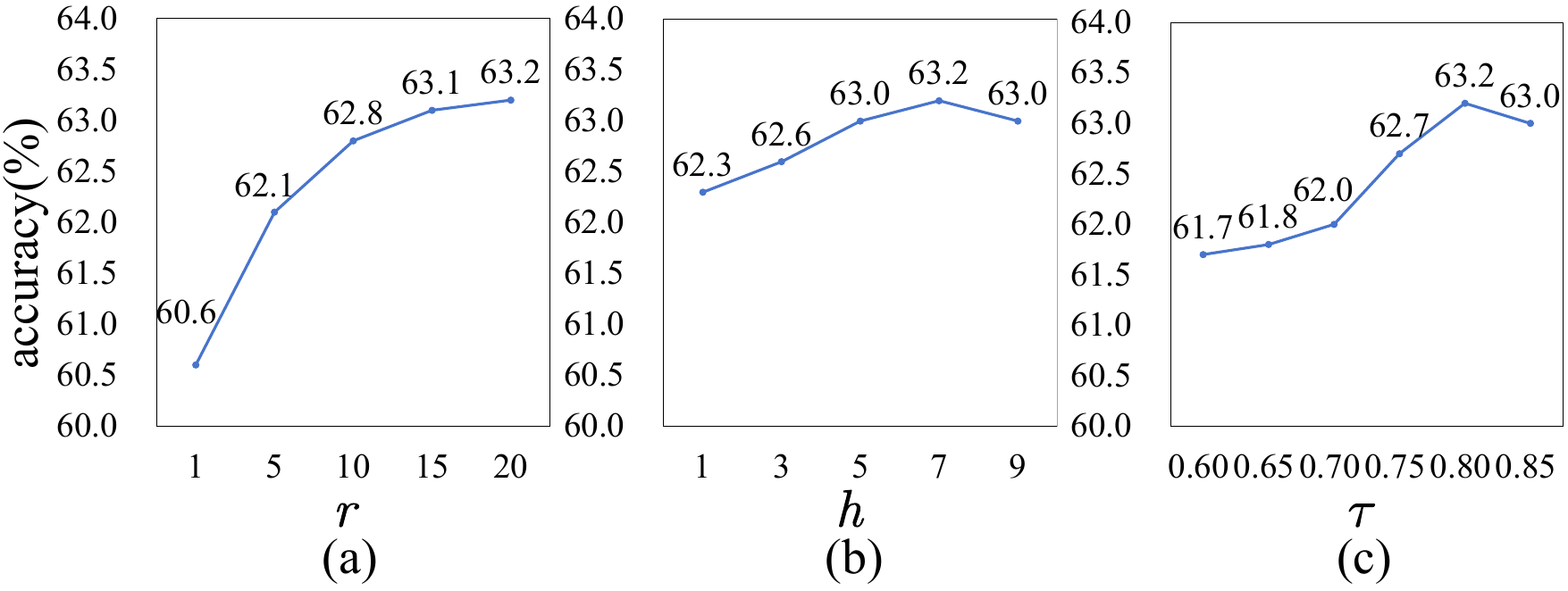}}
\caption{Hyper-parameter analysis on OK-VQA.}
\label{fig3}
\end{figure}

\begin{table}[t]
\centering
\caption{Comparison with the Visual Language Models on OK-VQA.}
\resizebox{1.0\linewidth}{!}{
\begin{tabular}{lccc|c}
\toprule
\textbf{Model} & MiniGPT-4 & InstructBLIP & Llama3.2-11B-Vision & KF-VQA\\ 
\midrule
\textbf{Acc} & 29.3 & 47.9 & 44.2 & \textbf{63.2}\\ 
\bottomrule
\end{tabular}
}
\label{tab4}
\end{table}

\begin{figure*}[t]
\centering
\centerline{\includegraphics[width=\textwidth]{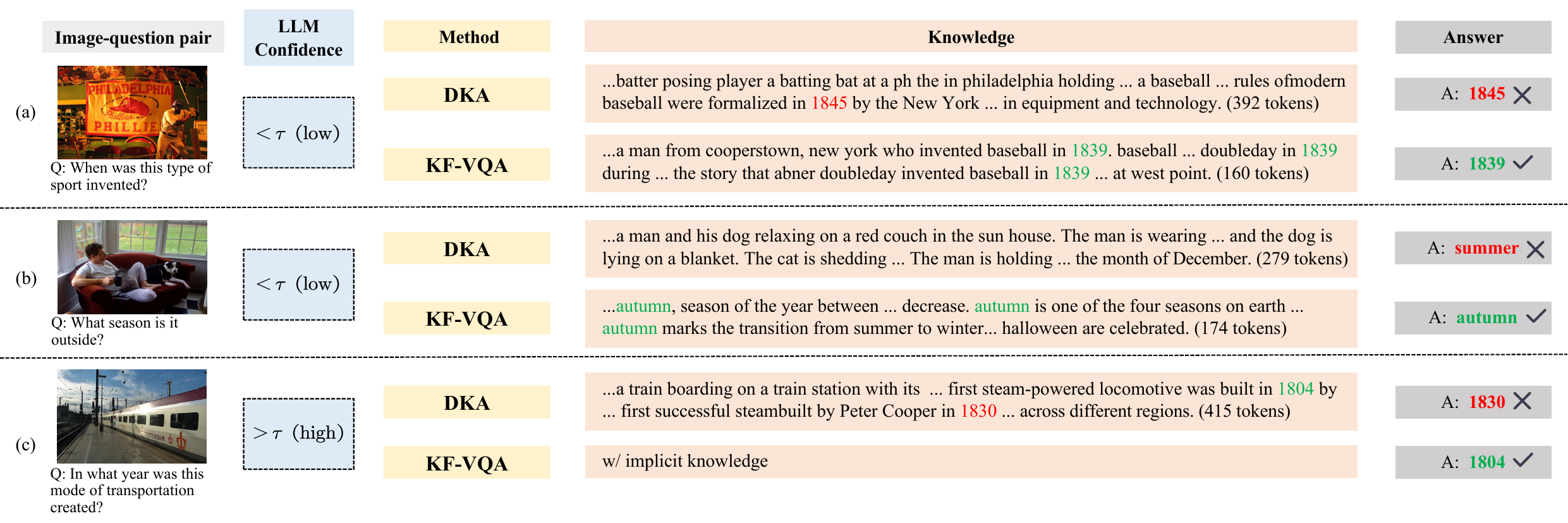}}
\caption{Case study on OK-VQA. Retrieved knowledge from different methods is listed along with the corresponding number of knowledge tokens.}
\label{fig4}
\end{figure*}

\subsection{Hyper-parameter Analysis}

\textbf{Effect of Retrieved Knowledge Numbers.} Fig.~\ref{fig3}(a) shows the effect of the number $r$ of retrieved knowledge documents on model prediction. The performance improves as the document number increases from 1 to 10, illustrating that more knowledge retrieved through our low-noise queries leads to improved reasoning capabilities. The model’s performance stabilizes when $r \geq 15$. This shows that useful knowledge segments may only originate from a small subset of documents, and simply increasing the number of retrieved documents does not result in sustained performance improvement.

\textbf{Effect of Knowledge Segment Numbers.} We evaluate the impact brought by the number $h$ of selected knowledge segments, with results shown in Fig.~\ref{fig3}(b). As $h$ increases from 1 to 7, the prediction accuracy improves incrementally, indicating that our knowledge filtering module effectively identifies useful segments from noisy documents, thereby enhancing the reasoning capabilities of LLMs. The model performance declines when $h > 7$, likely due to the inclusion of less relevant segments, which disrupt the model's prediction.

\textbf{Effect of Confidence Thresholds.} Fig.~\ref{fig3}(c) exhibits the effect of different threshold values of $\tau$. A low threshold $\tau$ means that LLMs cannot leverage knowledge even when answering questions with low confidence. As $\tau$ increases, the LLM could integrate more knowledge when it is uncertain about the prediction. The performance declines when $\tau > 0.8$. That is, if knowledge is introduced when LLMs could answer a question with high confidence, the irrelevant content from knowledge may degrade the model's performance, since the noise in knowledge cannot be fully filtered out.

\subsection{Comparison with Visual Language Models}
Since VLMs demonstrate strong abilities in visual recognition and language understanding, we compare our framework with VLMs, such as MiniGPT-4~\cite{zhu2023minigpt}, InstructBLIP~\cite{dai2023instructblip}, and Llama3.2-Vision~\cite{dubey2024llama}. The results in Table~\ref{tab4} show that our method outperforms these VLMs by a large margin. Although VLMs could better understand images, their knowledge reasoning abilities are inferior to that of LLMs. Our framework leverages the collaboration between the VLM and LLM to acquire high-quality knowledge that aids in prediction, providing an effective solution for KB-VQA.

\subsection{Qualitative Analysis} 
In Fig.~\ref{fig3}, we present a case study comparing KF-VQA with DKA. For cases (a) and (b), LLMs require external knowledge to answer the questions, as they could only make predictions with low confidence. The redundant knowledge retrieved by DKA hinders accurate predictions. Its retrieved texts, while abundant in visual information, are not only lengthy but also poorly relevant to the questions. In (a), the date ``1845'' misleads DKA into an incorrect prediction, whereas our concise knowledge contains essential clues. In (b), DKA fails to retrieve relevant knowledge. Our method focuses on key information, such as ``falling leaves'' from the image and ``season'' from the question, enabling the extraction of knowledge segments containing ``autumn''. Case (c) illustrates a scenario where LLMs could confidently answer the question without external knowledge. However, the knowledge retrieved by DKA includes noisy content, such as ``1830'', leading it 
to a wrong answer. In contrast, our selective knowledge integration strategy allows LLMs to leverage implicit knowledge for reasoning, thereby mitigating the impact of noisy knowledge.

\section{Conclusion}

This paper presents a training-free framework with knowledge focusing for KB-VQA, that mitigates the impact of knowledge noise by enhancing relevance and reducing redundancy. By identifying essential content from image-question pairs using VLMs, we construct low-noise queries to retrieve highly relevant knowledge. VLMs and LLMs further collaborate to filter out redundant segments from the obtained knowledge. Besides, for LLM reasoning, a selective knowledge integration strategy is designed to incorporate external knowledge only when the model lacks confidence in its answer, minimizing the impact of redundant knowledge on reasoning. Experimental results on benchmark datasets demonstrate that our method outperforms the state-of-the-art baselines.

\section*{Acknowledgments}
This work was supported by the National Natural Science Foundation of China (62406081), and the Guangxi Natural Science Foundation (No. 2025GXNSFBA069232).

\bibliographystyle{IEEEbib}
\bibliography{icme2025references}

\vspace{12pt}
\color{red}

\end{document}